\documentclass[10pt,twocolumn,letterpaper]{article}

\usepackage[pagenumbers]{cvpr} 
\usepackage{tabularx} 
\usepackage{booktabs}
\usepackage{array}
\usepackage{makecell}
\usepackage{tabularray}
\UseTblrLibrary{diagbox}
\usepackage{multirow}
\usepackage{color}
\usepackage{arydshln}

\usepackage{multirow} 
\usepackage{makecell} 

\definecolor{cvprblue}{rgb}{0.21,0.49,0.74}
\usepackage[pagebackref,breaklinks,colorlinks,citecolor=cvprblue]{hyperref}

\def\paperID{*****} 
\def\confName{CVPR}
\def\confYear{2024}

\title{Unifying Dimensions: A Linear Adaptive Approach to Lightweight Image Super-Resolution}

\author{Zhenyu Hu, Wanjie Sun\thanks{Corresponding author}\\
School of Remote Sensing and Information Engineering, Wuhan University\\ Wuhan 430079, China\\
\tt\small {\{zhenyuhu, sunwanjie\}@whu.edu.cn}
}
\begin{document}
\maketitle
\begin{abstract}
Window-based transformers have demonstrated outstanding performance in super-resolution tasks due to their adaptive modeling capabilities through local self-attention (SA). However, they exhibit higher computational complexity and inference latency than convolutional neural networks. In this paper, we first identify that the adaptability of the Transformers is derived from their adaptive spatial aggregation and advanced structural design, while their high latency results from the computational costs and memory layout transformations associated with the local SA. To simulate this aggregation approach, we propose an effective convolution-based linear focal separable attention (FSA), allowing for long-range dynamic modeling with linear complexity. Additionally, we introduce an effective dual-branch structure combined with an ultra-lightweight information exchange module (IEM) to enhance the aggregation of information by the Token Mixer. Finally, with respect to the structure, we modify the existing spatial-gate-based feedforward neural networks by incorporating a self-gate mechanism to preserve high-dimensional channel information, enabling the modeling of more complex relationships. With these advancements, we construct a convolution-based Transformer framework named the linear adaptive mixer network (LAMNet). Extensive experiments demonstrate that LAMNet achieves better performance than existing SA-based Transformer methods while maintaining the computational efficiency of convolutional neural networks, which can achieve a \(3\times\) speedup of inference time. The code will be publicly available at: \href{https://github.com/zononhzy/LAMNet}{https://github.com/zononhzy/LAMNet}.
\end{abstract}
\section{Introduction}
Single Image Super-Resolution (SISR) is a fundamental low-level task in computer vision that aims to recover realistic high resolution (HR) images from low resolution (LR) inputs. The primary goal is to reconstruct lost details and improve image quality. Thus, this technique is particularly crucial in applications that require high-quality images, such as medical imaging \cite{conf/cvpr/LiLTDWX022, journals/tip/CherukuriGSM20, journals/pami/HuRYCWMTZ22}, hyperspectral imagery \cite{journals/tgrs/LiGYW22}, and various other downstream tasks \cite{conf/cvpr/ShermeyerE19, conf/amdo/RastiUEA16}. This task is challenging because high-frequency information is often lost during degradation. Moreover, the uncertainty in mapping low-resolution images to high-resolution images makes the task ill-posed. To tackle this issue, many variants of Convolutional Neural Networks (CNN) \cite{conf/iccvw/YoonJYLK15, conf/eccv/ZhangLLWZF18, conf/cvpr/DaiCZXZ19, conf/mm/HuiGYW19, conf/cvpr/LiLCCGQD22, journals/tip/ZhangWJ19} and Vision Transformers (ViT) \cite{conf/cvpr/Chen000DLMX0021, conf/iccvw/LiangCSZGT21, conf/cvpr/Lu0LHZZ22, conf/iccv/li2023, conf/iccv/ZhouLGBCH23} have been proposed to model the nonlinear relationships between LR and HR image pairs. However, most related works \cite{conf/cvpr/LimSKNL17, conf/cvpr/ZhangTKZ018, conf/iccvw/LiangCSZGT21, conf/iccv/ZhouLGBCH23, conf/eccv/CondeCBT22} have focused on leveraging large models to obtain better learning capacity, hindering the application of super-resolution networks on practical scenarios. For SISR on resource-constrained devices, models must balance performance and computational cost. Consequently, both academia and industry are increasingly focused on developing lightweight super-resolution methods \cite{conf/mm/HuiGYW19, conf/cvpr/LiLCCGQD22, conf/nips/SunP022, conf/iccv/li2023, conf/eccv/ZhangZGZ22, conf/cvpr/WangCNLL23}, aiming to achieve good results with fewer parameters and lower computational costs. Currently, ViTs, with their efficient adaptive modeling capabilities, have shown significant performance gains over CNNs and are becoming increasingly dominant. As a result, many works focused on improving the multi-head self-attention (MHSA) mechanism and Transformer architecture for lightweight tasks to achieve better performance with lower computational costs.

Although these efficient ViT-based SR frameworks outperform CNN-based models with the same computational complexity, their runtime and training time are typically much longer than the latter. The main sources of inefficiency are identified by analyzing the runtime consumption of ViT-based models: 1) \textbf{Repeated memory layout modifications}: SR tasks focus on local texture patterns of images, leading current ViT-based SR models to use two primary operations to establish local relationships. First, input features are divided into non-overlapping patches for MHSA operation and then mapped back to the original plane \cite{conf/cvpr/Chen000DLMX0021, conf/iccvw/LiangCSZGT21}. Second, convolution operations are integrated into the framework \cite{conf/cvpr/FangLCZ22, conf/cvpr/WangCNLL23}. However, the dimension arrangements of convolution and Transformer operations differ, necessitating changes in memory layout. These operations do not increase computational complexity, but significantly slow down inference speed. 2) \textbf{Relative position encoding table}: The self-attention mechanism lacks the inductive bias of convolution for the image plane, meaning that it does not have positional priors. While the relative position encoding table helps the model capture local spatial structure information and patterns \cite{conf/iccv/LiuL00W0LG21}, its indexing and gradient backpropagation are inefficient. 3) \textbf{High computational complexity of the self-attention mechanism}: Although Local-ViT \cite{conf/iccvw/LiangCSZGT21, conf/iccv/ZhouLGBCH23} has mitigated this issue to some extent, it is still constrained by the patch size. 

Based on the above analysis, convolutional structures run more efficiently than Transformers and are better optimized with contemporary hardware accelerators. Inspired by previous work \cite{conf/cvpr/WangDCHLZHLLLWQ23, conf/iccv/li2023, conf/iclr/HanFDSC0022}, ViTs have two main advantages: 1) The multi-head self-attention (MHSA) mechanism's adaptive spatial aggregation capability allows for stronger and more robust representations at each position \cite{conf/iclr/DosovitskiyB0WZ21, conf/cvpr/ding2022scaling}, outperforming CNNs. 2) Advanced structural design: Transformers leverage layer normalization (LN) and feed-forward neural networks (FFN) \cite{conf/nips/VaswaniSPUJGKP17}, significantly boosting performance. Among them, the Deformable Convolution Network (DCN) \cite{conf/cvpr/WangDCHLZHLLLWQ23} uses convolutions to generate adaptive weights that simulate MHSA, and by incorporating advanced structures, they can outperform Transformers in complex tasks while avoiding the above drawbacks. However, the small convolution kernels limit the extraction of local features, and the computational burden from bilinear interpolation makes DCNs less suitable for lightweight tasks. 

\begin{figure}[!t]
    \centering
    \includegraphics[width=\columnwidth]{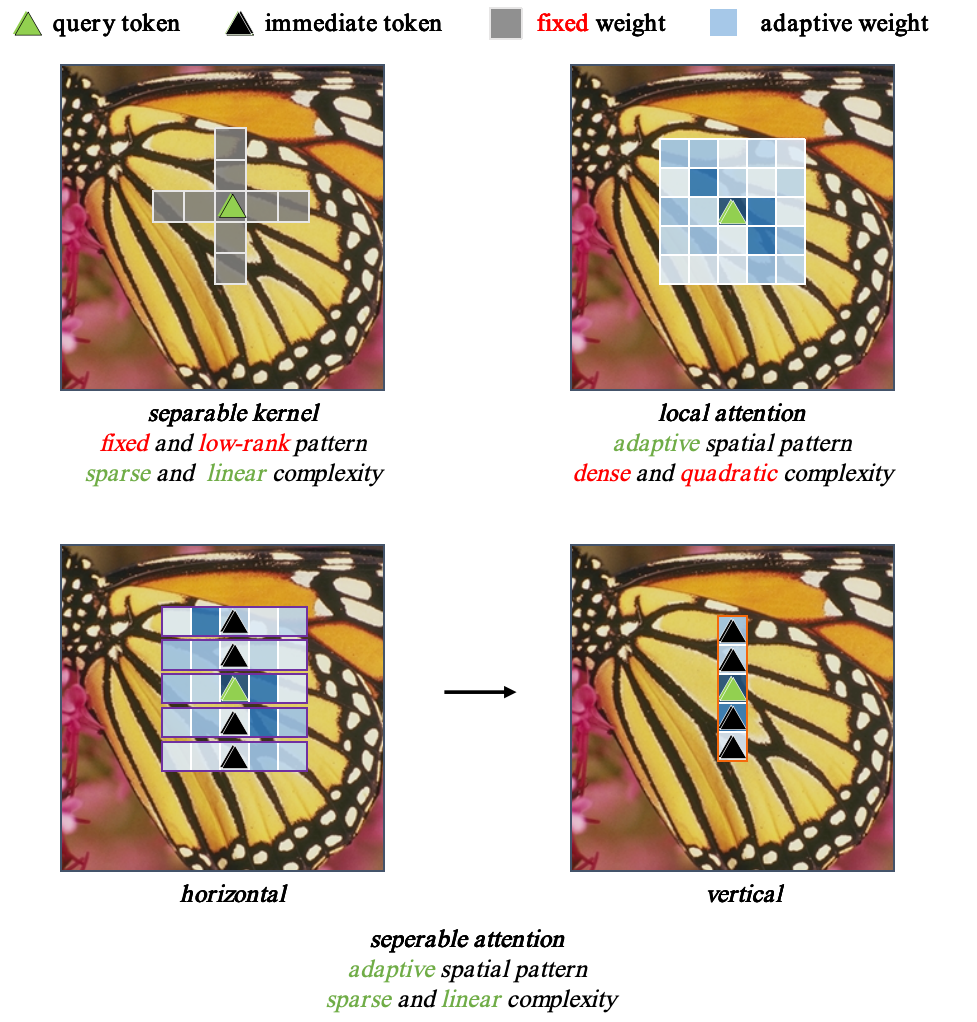}
    \caption{Comparison of different operators. Separable convolutions utilize one-dimensional kernels to achieve linear complexity in feature processing, which is inflexible. The local self-attention mechanism adaptively generates weights for query tokens, maintaining high computational complexity. The separable attention, while retaining its adaptive nature, linearly generates sparse weights to handle super-resolution tasks effectively.}
    \label{fig:attn_comp}
\end{figure}

In this work, we propose a lightweight convolution-based method with linear complexity for local-global adaptive modeling called the Linear-Spatial Adaptive Mixer (LSAM). As shown in Figure \ref{fig:attn_comp}, spatial separable convolution achieves a large receptive field with minimal computational cost. Still, it is limited to fixed patterns for aggregating information and is constrained by rank-1 weight matrices. In contrast, local attention mechanisms can adaptively adjust spatial patterns, but their complexity increases quadratically with window size. In super-resolution tasks, the positions of local features critical for reconstruction are often sparsely distributed \cite{conf/cvpr/MeiFZ21, journals/pami/SuGCYC23}. Therefore, we modified the spatial separable convolution design to deformable convolution to achieve a linear computational cost. By decomposing the adaptive spatial weights into sequential pixel-wise weights along both horizontal and vertical directions and incorporating the concept of visual localization \cite{conf/nips/YangLZDXYG21}, LSAM effectively sparsifies the 2D weight matrix while maintaining a high rank. We term this process Focal Separable Attention (FSA). However, several approaches \cite{conf/cvpr/WangCNLL23, conf/iccv/0014ZGKY023, conf/cvpr/ZamirA0HK022} have demonstrated that simultaneously considering both channel and spatial features can significantly enhance models' performance in low-level tasks. However, hybrid modeling often depends on varying memory layouts, and frequent changes in these layouts increase the model's latency without yielding performance improvements. To address this, we introduce an additional branch named the Channel Selective Mixer (CSM), designed for channel modeling and aligned with the memory layout of spatial operations. Furthermore, we propose a parameter-free Information Exchange Module (IEM) to facilitate efficient interaction between the two branches, characterized by its linear complexity.

From the perspective of structural design, while the use of spatial gates has been employed to improve the spatial modeling capability of FFN adapted for low-level tasks \cite{conf/iccv/0014ZGKY023, conf/cvpr/ZamirA0HK022}, this gating mechanism inadvertently diminishes the capacity of the channel dimension. To mitigate this issue, we design a Dual-Gated Feed-Forward Network (DGFN), which simultaneously applies the spatial-gate operation to self and another branch, ensuring that the gating process preserves the diversity of channel features.

Based on the aforementioned module design, we integrated it with the superior structure of the Transformer to create an efficient Linear Adaptive Mixer Network (LAMNet) for SR. The proposed LAMNet offers shorter inference time at the same model size while delivering superior performance and visual quality. In summary, the main contributions of this paper are as follows: 
\begin{itemize}
\item{We propose the Unified Linear Mixer (ULM), comprising both spatial and channel branches. The spatial branch adopts the Linear-Spatial Adaptive Mixer (LSAM), which equips convolution operations with the adaptive modeling capability of MHSA and leverages the sparse design of separable convolutions. The channel branch employs the Channel Selective Mixer (CSM) to address the limitations in channel dimension modeling.}
\item{We develop an efficient parameter-free Information Exchange Module (IEM) that enables the sharing of both spatial and channel information between the two branches, thereby addressing the issue of their mutual independence.}
\item{We introduce the Dual-Gated Feed-Forward Network (DGFN), which not only strengthens the spatial gating capabilities but also enhances the information capacity within the channel dimension.}
\item{We propose an efficient Linear Adaptive Mixer Network (LAMNet), combining the advanced structural design of the ViT with the inference efficiency of CNN, striking a balance among computational cost, latency, and model performance.}
\end{itemize}

The remainder of this paper is organized as follows: Section \ref{sec:realted} reviews convolutional neural networks and transformer-based SR networks. Section \ref{sec:method} presents the proposed LAMNet and details the processing flow of its core components. Section \ref{sec:experiments} evaluates the model’s performance both quantitatively and qualitatively and conducts ablation studies on its various components. Finally, we conclude the paper in Section \ref{sec:conclusion}.
\section{Related work}
\label{sec:realted}
In this section, we review the representative CNN-based and Transformer-based approaches.

\subsection{Lightweight SISR Model}
In recent years, neural networks' powerful learning capabilities have driven the development of many effective SISR methods. Dong et al. \cite{journals/pami/DongLHT16} proposed the groundbreaking SRCNN, a three-layer CNN that can directly model the mapping from LR to HR. Subsequently, Kim et al. \cite{conf/cvpr/KimLL16, conf/cvpr/KimLL16a} introduced VDSR and DRCN, enhancing accuracy through global residual learning and recursive layers. Ledig et al. \cite{conf/cvpr/LedigTHCCAATTWS17} developed SRResNet, which achieves super-resolution of LR images by combining adversarial learning with 16 residual blocks. The NTIRE 2017 super-resolution challenge \cite{conf/cvpr/TimofteAG0ZLSKN17} winner EDSR \cite{conf/cvpr/LimSKNL17} improved upon SRResNet by removing batch normalization and expanding the network structure, thereby enhancing quantitative metrics and visual quality. Subsequently, RDN \cite{conf/cvpr/ZhangTKZ018} and RCAN \cite{conf/eccv/ZhangLLWZF18} respectively increased the network depth to over 100 and 400 layers, surpassing EDSR. Recently, Transformer-based super-resolution models have demonstrated superior performance, such as SwinIR \cite{conf/iccvw/LiangCSZGT21}. SwinIR, built based on Shifted Window Multi-Head Self-Attention ((S)W-MSA), uses a three-stage framework to improve efficiency. Building on SwinIR, Chen et al. \cite{conf/cvpr/ChenWZ0D23} proposed the Hybrid Attention Transformer (HAT), which combines channel attention with window-based self-attention to achieve state-of-the-art results. Many Transformer-based networks \cite{conf/iccv/ZhouLGBCH23, conf/eccv/CondeCBT22} have proven their excellent performance by incorporating (S)W-MSA.

However, due to the high computational cost, most methods are limited in their applicability to real-world scenarios. Several lightweight and efficient SISR methods using novel model architectures have been proposed to address this issue. For example, IDN \cite{conf/cvpr/HuiWG18} employs an information distillation network to fuse features selectively; IMDN \cite{conf/mm/HuiGYW19} improves upon this to create a lighter and faster model. Building on IMDN, Liu et al. \cite{conf/mm/HuiGYW19} proposed the Residual Feature Distillation Network (RFDN), which combines feature distillation connections and shallow residual blocks to achieve better performance with reduced computational cost. Later, BSRN \cite{conf/cvpr/LiLCCGQD22} used the same feature extraction structure as IMDN and introduced Blueprint Separable Convolutions (BSConv) to replace standard convolutions. These convolution-based networks have achieved excellent results in the NTIRE and AIM lightweight super-resolution challenges, particularly in latency, due to their consistent and efficient convolution operations, providing better modularity and reproducibility. In contrast, the adaptive aggregation capability of MHSA is more suited to SR tasks, leading to an increased research focus on developing lightweight Transformer networks. Zhang et al. \cite{conf/eccv/ZhangZGZ22} proposed the Efficient Long-range Attention Network (ELAN), which uses a shared state mechanism to accelerate SR tasks. It has also become common to combine the local feature extraction ability of convolutions with the high-frequency extraction ability of Transformers to enhance their expressive power. For instance, the Hybrid Network of CNN and Transformer (HNCT) \cite{conf/cvpr/FangLCZ22} combines (S)W-MSA layers and convolution-based enhanced spatial attention blocks for SR tasks. Chen et al. \cite{conf/iccv/0014ZGKY023} proposed the Dual Aggregation Transformer (DAT), which features adaptive interaction modules that exchange spatial and channel information in the convolution and attention branches. However, these methods often have much higher actual inference times than convolutional networks of the same scale due to frequent memory layout changes, high computational complexity, and frequent memory access associated with the self-attention mechanism. Although DLGSANet \cite{conf/iccv/li2023} attempts to simulate the pixel-wise adaptive aggregation capability of MHSA using convolutions, It continues to use the test-time local converter (TCL) \cite{conf/eccv/ChuCCL22} method, which repeatedly alters the memory layout. Based on the methods of DLGSANet and DCN \cite{conf/cvpr/WangDCHLZHLLLWQ23}, this paper designs the Linear-Spatial Adaptive Mixer (LSAM), combining the efficiency of convolutional operations with the adaptive modeling capabilities of Transformers, performing sparse modeling of local areas with linear time complexity related to window size. Simultaneously, the other Channel Selective Mixer (CSM) branch interacts with LSAM via the efficient Information Exchange Module (IEM), seamlessly integrating spatial and channel information.

\subsection{Feed-Forward Network}
In the Transformer architecture, the FFN is responsible for channel dimension feature transformation, enhancement, and nonlinear modeling, which improves its performance in complex tasks \cite{conf/nips/VaswaniSPUJGKP17}. However, the original FFN overlooks the significance of spatial information in low-level tasks, leading to excessive computational costs dedicated to channel expansion. As a result, various approaches have been developed to introduce spatial information into FFN, aiming to boost its spatial modeling capabilities. For instance, Chen et al. \cite{conf/iccv/0014ZGKY023} and Zamir et al. \cite{conf/cvpr/ZamirA0HK022} incorporate additional nonlinear spatial information and mitigate channel redundancy through spatial gating operations. However, these spatial gating methods reduce the channel dimension, impairing the FFN's ability to capture relationships among high-dimensional channel features. To address this issue, we propose the Dual-Gated Feed-Forward Network (DGFN), compensating for the reduction and enhancing spatial feature modeling by introducing a self-gate operation.
\section{Methodology}
\label{sec:method}

\subsection{Linear Adaptive Mixer Network}
\begin{figure*}[ht]
    \centering
    \includegraphics[width=\textwidth]{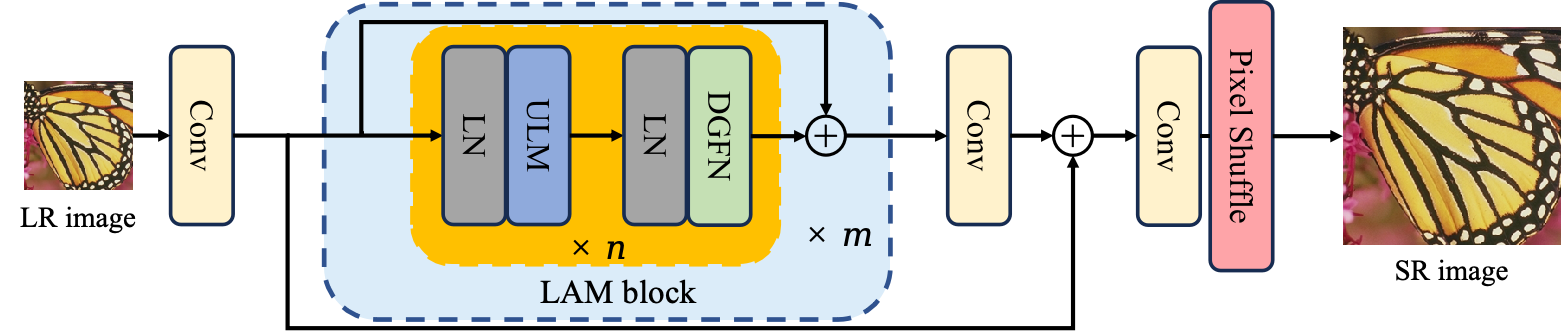}
    \caption{The overall architecture of the proposed Linear Adaptive Mixer Network (LAMNet) is presented, where the core operation, highlighted in the blue area, is the Linear Adaptive Mixer (LAM) Block. The Unified Linear Mixer (ULM), Dual-Gated Feed-Forward Network (DGFN), and LayerNorm constitute the basic Transformer block, which is stacked \(n\) times to form the main structure of the LAM block. Features are processed within the model to optimize efficiency using the \(C\times H\times W\) memory layout.}
    \label{fig:overall}
\end{figure*}
To combine the efficiency of convolution operations with the adaptive modeling capabilities of Transformer architectures, we design an efficient Linear Adaptive Mixer Network (LAMNet) for SISR. As illustrated in Figure \ref{fig:overall}, LAMNet is based primarily on the SwinIR \cite{conf/iccvw/LiangCSZGT21} framework, comprising shallow feature extraction, deep feature extraction, and image reconstruction modules. In the SwinIR framework, features are represented with \(H\times W\times C\) layout, which is not optimal for convolution operations and spatial feature extraction. Therefore, we utilize the \(C\times H\times W\) layout, which is the default memory arrangement for convolution operations. The input and output of LAMNet are defined as \(\boldsymbol{I}_{LR}\) and \(\boldsymbol{I}_{HR}\), respectively. Initially, the input image undergoes a rapid dimensional expansion through a convolution layer to facilitate deeper processing, 
\begin{equation}
    \boldsymbol{X}_{shallow} = \mathrm{F}_{SF}(\boldsymbol{I}_{LR}),
\end{equation}
where the \(\mathrm{F}_{SF}(\cdot)\) layer performs both shallow feature extraction and channel expansion, \(\boldsymbol{X}_{shallow}\) is the shallow features. Subsequently, these shallow features are fed into multiple consecutive Linear Adaptive Mixer (LAM) blocks for one-dimensional feature extraction across both channel and spatial dimensions. Each LAM block comprises several Unified Linear Mixers (ULM) and Dual-Gated Feed-Forward Networks (DGFN), which collectively constitute the components of a standard Transformer block. The complete operation of each LAM block can be expressed as follows: 
\begin{align}
\begin{split}
    \boldsymbol{X}_{lam}^{i} &= \mathrm{F}_{LAM}(\boldsymbol{X}_{in}) \\ &= (\mathrm{F}_{G2FN}\mathrm{F}_{ULM})^{1\dots n}(\boldsymbol{X}_{in}) + \boldsymbol{X}_{in},
\end{split}
\end{align}
where \(\mathrm{F}_{LAM}(\cdot)\), \(\mathrm{F}_{G2FN}(\cdot)\) and \(\mathrm{F}_{ULM}(\cdot)\) denote each module in the LAM block; \((\mathrm{F}_{G2FN}\mathrm{F}_{ULM})^{1\dots M}\) indicates that G2FN and ULM cross-stack \(n\) times, forming a common Transformer block. \(\boldsymbol{X}_{lam}^{i}\) and \(\boldsymbol{X}_{in}\) are the input and output of the \(i\)-th LAM block after the operation of \(m\) blocks; thus,we can obtain the final deep features \(\boldsymbol{X}_{deep}\)
\begin{equation}
    \boldsymbol{X}_{deep} = \mathrm{Conv}({\mathrm{F}_{LAM}^{i\cdots m}(\boldsymbol{X}_{shallow})}) + \boldsymbol{X}_{shallow}.
\end{equation}
Here, \(\mathrm{F}_{LAM}^{i\cdots m}(\cdot)\) and \(\mathrm{Conv}(\cdot)\) represent \(m\) consecutive LAM blocks and a \(3\times 3\) convolution, respectively. Finally, to obtain the SR image, we use a reconstruction layer to upsample the deep features
\begin{equation}
    \boldsymbol{I}_{SR} = \mathrm{F}_{REC}(\boldsymbol{X}_{deep}),
\end{equation}
where \(\mathrm{F}_{REC}(\cdot)\) contains a convolution and pixelshuffle \cite{conf/cvpr/ShiCHTABRW16} layer.
Given a training dataset \(\{\boldsymbol{I}_{LR,n},\boldsymbol{I}_{HR,n}\}_{n=1}^{N}\) with \(N\) ground-truth images \(\boldsymbol{I}_{HR}\), and their corresponding LR counterpart \(\boldsymbol{I}_{LR}\), we employ L1 loss to optimize the parameters of the proposed model:
\begin{equation}
    \begin{aligned}\mathrm{L}\left(\Theta\right)&=\frac{1}{N}\sum_{n=1}^{N}\left\|\boldsymbol{I}_{SR,n}- \boldsymbol{I}_{HR,n}\right\|_{1}\,,\end{aligned}
\end{equation}
where \(\Theta\) is the model parameters.

Previous research works \cite{conf/cvpr/ZamirA0HK022, conf/cvpr/ChenWZ0D23} indicate that Transformers outperform convolution-based networks in low-level tasks when operating at the same computational complexity. This advantage is largely attributed to the superior network design of Transformers and their Multi-Head Self-Attention (MHSA) mechanism. MHSA, a crucial component, allows the model to process multiple positions in the input sequence simultaneously, thereby capturing long-range dependencies more effectively. This mechanism enables query tokens to adjust the aggregation weights according to similarity criteria. Given the input flattened feature maps \(\boldsymbol{X}\in \mathbb{R}^{N\times C}\), three linear layers are applied to attain query \(\boldsymbol{Q}\), key \(\boldsymbol{K}\), and value \(\boldsymbol{V}\) embeddings respectively. Then, the attention \(\boldsymbol{attn}_i\) of the query token \(\boldsymbol{q}_i\) in \(\boldsymbol{Q}\) can be generally formulated as
\begin{equation}
    \boldsymbol{attn}_i = \sum_{j=1}^{N}\frac{\mathrm{Sim}(\boldsymbol{q}_i, \boldsymbol{k}_j)}{\sum_{l=1}^{N}(\mathrm{Sim}(\boldsymbol{q}_i, \boldsymbol{k}_l))}\boldsymbol{v}_j,
\end{equation}
where \(\mathrm{Sim}(\boldsymbol{q}_i, \boldsymbol{k}_j)\) measures the similarity between \(\boldsymbol{q}_i\) and \(\boldsymbol{k}_j\), and generates dynamic weights based on normalization. The MHSA mechanism typically employs exponential similarity functions to emphasize the weights of highly similar tokens, resulting in a sparse weight matrix for sequences or images. Transformer was originally designed for natural language processing tasks. When applied to low-level tasks, it often serializes two-dimensional data, disregarding the regular structure of images and frequently altering memory layouts. The computational intensity of the MHSA mechanism increases quadratically with the window size, resulting in higher inference latency for lightweight tasks. Therefore, we aim to harness the adaptive capabilities of MHSA using convolution. This analysis indicates that the key to this adaptiveness is the generation of dynamic weights. Previous works have applied solutions \cite{conf/cvpr/WangDCHLZHLLLWQ23, conf/iclr/HanFDSC0022} from high-level tasks to SR tasks \cite{conf/iccv/li2023}. However, these approaches often require a significant amount of parameters and computational resources for dynamic weight generation and fail to account for the sparse nature of dynamic weight matrices in SR tasks.

Therefore, we aim to develop a method for predicting dynamic weights using a network, reducing the computational costs of weight generation by leveraging sparsity. We introduce the Unified Linear Mixer (ULM), designed to achieve this by using spatially separable convolutions in the spatial branch. As depicted in the Figure \ref{fig:ulm}, the input feature \(\boldsymbol{x}\) first passes through a \(1\times 1\) convolution in the ULM to produce a mixed feature \(\boldsymbol{X}\), similar to the token generation in MHSA. Feature \(\boldsymbol{X}\) is then processed through the spatial and channel branches to gather information across different dimensions.

\subsection{Unified Linear Mixer}
\begin{figure*}[ht]
    \centering
    \includegraphics[width=0.9\textwidth]{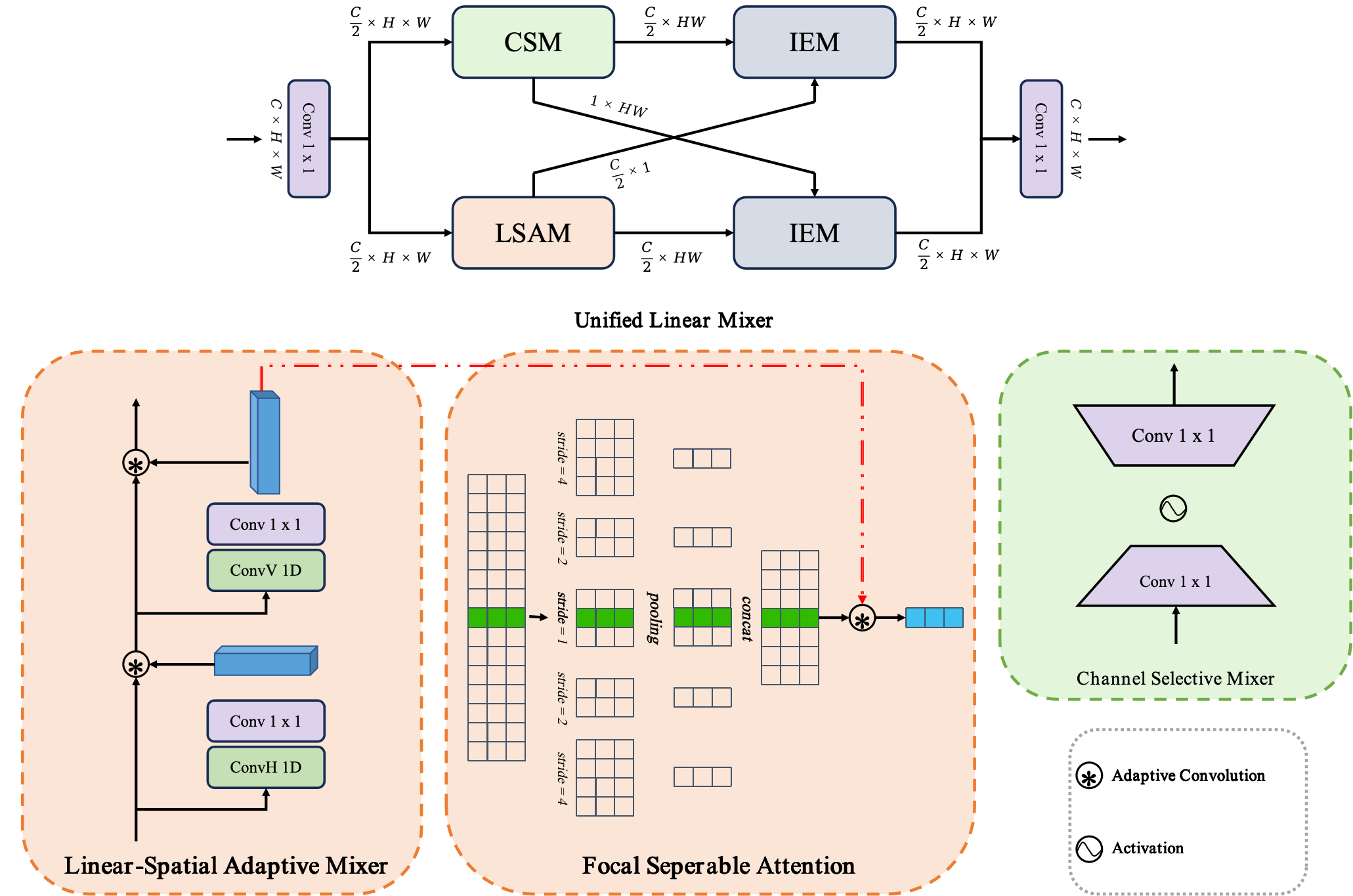}
    \caption{The framework of Unified Linear Mixer (ULM), which incorporates Linear-Spatial Adaptive mixer (LSAM) for spatial components and Channel Selective Mixer (CSM) for channel components. Spatial information is then continuously aggregated in horizontal and vertical directions using the Focal Seperable Attention (FSA). Subsequently, the two Information Exchange Modules facilitate information interaction between the spatial and channel branches.}
    \label{fig:ulm}
\end{figure*}

\textbf{Spatial Branch.} The spatial branch primarily consists of a Linear-Spatial Adaptive mixer (LSAM), where a convolution block predicts dynamic weights for spatial regions. This block typically shares the same receptive field as the subsequent dynamic convolution, ensuring stability in weight generation. Although our linear mixer reduces two-dimensional weights to one dimension, the kernel's local coverage remains large, maintaining high computational complexity if a single two-dimensional depth-wise convolution (DWConv) is used. Therefore, we employ one-dimensional convolution independently for each direction to generate the corresponding dynamic weights.

Additionally, we propose a Focal strategy for the local token mixer from a visual probability perspective to minimize the cost of modeling sparsity. This enlarges the information capture area while reducing the weight generation cost. For SR tasks, both local and global operations are used to capture token information that is critical for reconstruction. Dense local information is typically modeled with token-by-token weights, fundamental to convolutional neural networks. On a global scale, images often contain regions with similar textures but different scales, which can guide reconstruction. However, these textures are usually confined to small local areas. Standard operations treat all tokens within the receptive field equally, increasing feature aggregation cost as the local receptive field grows. Thus, we design a method to replace distant regions with a single token. Figure \ref{fig:ulm} shows that features are averaged in the vertical directions with different strides and then convolved with learned dynamic weights, effectively mimicking MHSA operations in the spatial domain. Each square cell represents a visual token from the original feature map or a pooled summarized token.
Suppose we have a \(15\times 3\) vertical window. For the center token on the plane, represented by the green token in the figure, the window is divided into sub-windows based on predefined strides of \([1, 2, 4]\) and their corresponding steps of \([1, 1, 1]\). Pooling is applied to each sub-window using different strides to generate agent tokens. These agent tokens are concatenated in spatial order and convolved with dynamic weights to produce the output token. We call this process Focal Seperable Attention (FSA). To enhance the model's ability to capture diverse features and thus improve its expressiveness, we incorporate the concept of MHSA, generating distinct weights for each group. Upon embedding FSA, the overall procedure of LSAM can be described as follows:
\begin{equation}
\begin{aligned}
    \boldsymbol{W}_{H} &= \mathrm{Conv}(\mathrm{ConvH}(\boldsymbol{X})),\\
    \boldsymbol{X}_{H} &= \mathrm{f}_{H}(\boldsymbol{X}, \boldsymbol{W}_H),\\
    \boldsymbol{W}_{V} &= \mathrm{Conv}(\mathrm{ConvV}(\boldsymbol{X}_{H})),\\
    \boldsymbol{X}_{s} &= \mathrm{f}_{V}(\boldsymbol{X}_{H}, \boldsymbol{W}_V), 
\end{aligned}
\end{equation}
where \(\mathrm{f}_{H}(\cdot, \boldsymbol{W}_{H})\) and \(\mathrm{f}_{H}(\cdot, \boldsymbol{W}_{V})\) use the dynamic weight \(\boldsymbol{W}_{H}, \boldsymbol{W}_{V}\) to blend tokens for input features in horizontal and vertical directions, respectively.

\textbf{Channel Branch} Numerous studies \cite{conf/cvpr/WangCNLL23, conf/iccv/0014ZGKY023, conf/cvpr/ZamirA0HK022} have shown that the interaction between channel and spatial dimensions facilitates deeper feature extraction, thereby enhancing overall performance in super-resolution tasks. To this end, we propose the Channel Selective Mixer (CSM) module, which selectively filters out irrelevant information by compressing features along the channel dimension, as illustrated in Figure \ref{fig:ulm}. The CSM can be formulated as follows:
\begin{equation}
    \boldsymbol{X}_{c} = \mathrm{Exp}(\mathrm{Relu}(\mathrm{Sqz}(\boldsymbol{X}))),
\end{equation}
where \(\mathrm{Exp}\) and \(\mathrm{Sqz}\) represent the channel expansion and squeeze operations, respectively, both of which are implemented using a single \(1\times1\) convolution.

\textbf{Information Exchange Module} While the two branches independently aggregate information across spatial and channel dimensions, simply concatenating features results in isolated interactions between the two types of features. Common approaches to facilitate feature interaction involve attention mechanisms \cite{journals/corr/abs-2407-05878} or convolutional attention mechanisms \cite{conf/iccv/0014ZGKY023}. However, the former is computationally intensive and involves high parameters, making them unsuitable for lightweight models. To address this, we propose an efficient, parameter-free Information Exchange Module that enables rapid information fusion between the two branches. Specifically, the channel branch enhances its spatial features using statistical features from the spatial branch, while the spatial branch applies channel attention using statistical features from the channel branch. For the spatial branch, we compute the mean of the other branch’s features along the channel dimension to generate a single query token, with the spatial branch’s features serving as key and value tokens. Channel attention is then achieved by calculating similarity. The entire process can be described as follows:
\begin{equation}
\begin{aligned}
    \boldsymbol{query}_{c} &= \sum_{i=1}^{\frac{C}{2}} \boldsymbol{X}_{c}^{i}, \\
    \boldsymbol{key}_{s}, \boldsymbol{key}_{s} &= \boldsymbol{X}_{s}, \boldsymbol{X}_{s}, \\
    \boldsymbol{X}_{s}^{IEM} &= \mathrm{Sigmoid}(\frac{\boldsymbol{query}_{c} \times \boldsymbol{key}_{s}^{T}}{H\times W}) \cdot \boldsymbol{value}_{s}.
\end{aligned}
\end{equation}
Here, \(\boldsymbol{X}_{c}, \boldsymbol{X}_{s} \in \mathbb{R}^{\frac{C}{2}\times H W}\), the \(\mathrm{Sigmoid}\) function serves as the similarity metric. Similarly, for the channel branch, we average the other branch’s spatial features to use as the query token, with the channel features serving as the key and value tokens to achieve spatial enhancement. It can be formulated as follows:
\begin{equation}
\begin{aligned}
    \boldsymbol{query}_{s} &= \sum_{i=1}^{H\times W} \boldsymbol{X}_{s}^{i}, \\
    \boldsymbol{key}_{c}, \boldsymbol{key}_{c} &= \boldsymbol{X}_{c}, \boldsymbol{X}_{c}, \\
    \boldsymbol{X}_{c}^{IEM} &= \mathrm{Sigmoid}(\frac{\boldsymbol{query}_{s} \times \boldsymbol{key}_{c}^{T}}{\frac{C}{2}}) \cdot \boldsymbol{value}_{c},
\end{aligned}
\end{equation}
Due to the IEM having only a single query token, the overall computational cost is minimized, equivalent to the complexity of two dot product operations.

\subsection{Dual-Group Feed-Forward Network}
\begin{figure}[!t]
    \centering
    \includegraphics[width=0.9\columnwidth]{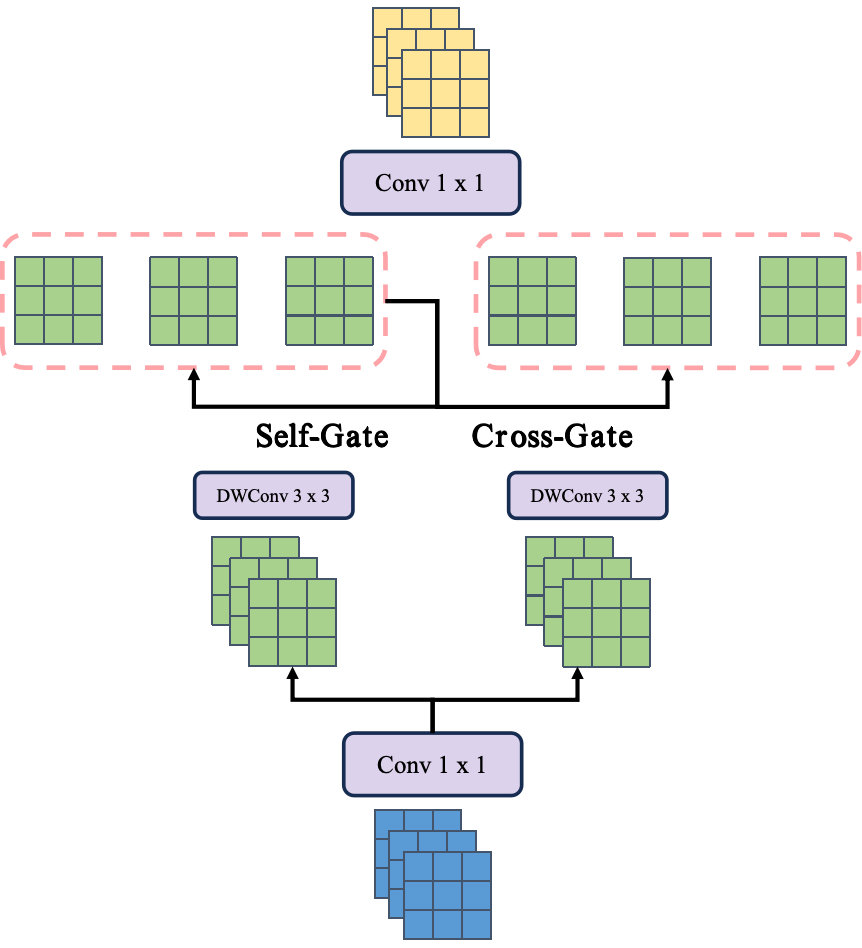}
    \caption{The overall structure of the Dual-Gated Feed-Forward Network includes a self-gate and cross-gate operation to enhance the channel dimension and mitigate the gate's impact on it.}
    \label{fig:dgfn}
\end{figure}
The Feed-Forward Network is crucial for the Transformer architecture's feature transformation and nonlinear mapping \cite{conf/nips/VaswaniSPUJGKP17}. It typically comprises two linear transformations followed by a nonlinear activation function, aiming to enhance the model's expressive power while keeping the input and output dimensions consistent. By processing features independently at each position, the Feed-Forward Network captures more complex feature relationships, complementing the information extracted by the self-attention mechanism. This network enhances the Transformer's ability to represent multi-level features, leading to a deeper understanding of the input features.

However, the FFN tends to focus too much computation on the channel dimension, overlooking the importance of local spatial features in super-resolution tasks. To address this, many approaches have sought to simplify the FFN by incorporating depthwise convolution (DWConv) after channel expansion to enhance spatial information, along with spatial gating to improve spatial feature representation. Nevertheless, this gating operation reduces the channel dimension, limiting the FFN's capacity to explore complex relationships in high-dimensional space. To address this issue, we introduce a self-gate into the spatial-gate process to recover the channel features lost during gating. The self-gate also ensures alignment with the cross-gate operation applied to the other branch, as shown in Figure \ref{fig:dgfn}. The process can be expressed as follows:
\begin{equation}
\begin{aligned}
    \boldsymbol{X}_{exp1} &= \mathrm{Exp1}(\boldsymbol{X}), \\
    \boldsymbol{X}_{exp2} &= \mathrm{DWConv}(\boldsymbol{X}_{exp1}), \\
    \boldsymbol{X}_{1}, \boldsymbol{X}_{2} &= \mathrm{Split}(\boldsymbol{X}_{exp2}), \\
    \boldsymbol{X}_{s-gate} &= \boldsymbol{X}_{1} \cdot \mathrm{Gelu}(\boldsymbol{X}_{1}) \\
    \boldsymbol{X}_{c-gate} &= \boldsymbol{X}_{2} \cdot \mathrm{Gelu}(\boldsymbol{X}_{1}), \\
    \boldsymbol{X'} &= \mathrm{Sqz}(\mathrm{Cat}(\boldsymbol{X}_{s-gate}, \boldsymbol{X}_{c-gate}), 
\end{aligned}
\end{equation}
Here, \(\boldsymbol{X}, \boldsymbol{X'} \in \mathbb{R}^{C\times H \times W}\) represent the input and output features, \(\boldsymbol{X}_{(\cdot)}\) denotes the intermediate feature. \(\mathrm{Exp1}\) and \(\mathrm{Sqz}\) are the channel expansion and
squeeze operations, respectively. \(\mathrm{DWConv}\) is implemented by \(3\times3\) DWConv for spatial information extraction.

\subsection{Complexity Analysis of ULM and DGFN}
\label{sec:cost_anaysis}
\begin{table*}[htb]
\centering
\caption{Parameter comparison among SwinIR, DLGSANet, and LAMNet}
\small
\label{tbl:params_cp}
\renewcommand{\arraystretch}{1.3}
\begin{tblr}{
  cells = {c},
  vline{2-4} = {-}{},
  hline{1,5} = {-}{0.08em},
  hline{2} = {-}{0.05em},
}
\diagbox{Part}{Model} & SwinIR      & DLGSANet                                 & LAMNet                         \\
Token Mixer              & $4C^2+K^4$ & $\frac{5}{2}C^2+\frac{G+1}{2}K^2C$      & $\frac{5}{2}C^2+(G+1)KC$      \\
FFN                      & $4C^2$     & $3C^2+18C$                              & $4C^2+18C$                    \\
Total                    & $8C^2+K^4$ & $\frac{11}{2}C^2+18C+\frac{G+1}{2}K^2C$ & $\frac{13}{2}C^2+18C+(G+1)KC$ 
\end{tblr}
\end{table*}

\begin{table*}[htb]
\centering
\caption{Flops comparison among SwinIR, DLGSANet, and LAMNet}
\small
\label{tbl:flops_cp}
\renewcommand{\arraystretch}{1.3}
\resizebox{\textwidth}{!}{
\begin{tblr}{
  cells = {c},
  vline{2-4} = {-}{},
  hline{1,5} = {-}{0.08em},
  hline{2} = {-}{0.05em},
}
\diagbox{Part}{Model} & SwinIR                & DLGSANet                                       & LAMNet                               \\
Token Mixer              & $4HWC^2+2HWK^2$      & $\frac{5}{2}HWC^2+\frac{G+3}{2}K^2HWC$        & $\frac{5}{2}HWC^2+(G+2)HWKC$        \\
FFN                      & $4HWC^2+2HWC$        & $3HWC^2+20HWC$                                & $4HWC^2+21HWC$                      \\
Total                    & $8HWC^2+(2K^2+2)HWC$ & $\frac{11}{2}HWC^2+20HWC+\frac{G+3}{2}K^2HWC$ & $\frac{13}{2}HWC^2+21HWC+(G+2)KHWC$
\end{tblr}
}
\end{table*}
To better demonstrate the complexity and parameter variations of our proposed token mixer (ULM) and FFN (DGFN), we compare them with the standard local Transformer network SwinIR \cite{conf/iccvw/LiangCSZGT21} and the dynamic-conv-based super-resolution network, DLGSANet \cite{conf/iccv/li2023}. Let \(H\), \(W\), and \(C\) represent the height, width, and channel number of the input features, while \(K\) stands for the window size—referring to the window size in SwinIR and the convolution kernel size in LAMNet and DLGSANet. Note that in LAMNet, the FSA pooling operation reduces the convolution kernel size compared to the window size. \(G\) denotes the number of channel groups, which enhances the model's ability to capture multi-scale and multi-level features. The comparison of parameters and FLOPs among the three models is presented in Tables \ref{tbl:params_cp} and \ref{tbl:flops_cp}. Specifically, we compare three parts: Token Mixer, FFN, and their combined complexity, while excluding shared components like LayerNorm and residual connections. The comparison is conducted under the same lightweight network settings, where \(H\), \(W\), \(C\), \(K\), and \(G\) are set to 1280, 720, 64, 8, and 64, respectively. Currently, LAMNet has 30K parameters, fewer than SwinIR's 37K and DLGSANet's 34K. SwinIR exhibits the highest FLOPs at 38G, followed by LAMNet at 26G and DLGSANet at 22G. When calculating the complexity-to-parameters ratio, SwinIR achieves the highest value, while dynamic-conv-based approaches show relatively lower values, largely due to the computationally intensive yet parameter-free nature of MHSA. In other words, SA-based Transformer models tend to have an advantage in terms of parameters compared to conv-based models with similar computational complexity. However, during inference, the memory usage of parameters is minimal compared to features, adding only a slight storage burden. Therefore, when comparing these methods, the focus is more on inference time.
\section{Experiments}
\label{sec:experiments}
In this section, we provide relevant experimental details, descriptions, and results to verify the proposed LAMNet's effectiveness and excellence.

\subsection{Experimental Settings}
\begin{table*}
\caption{Quantitative comparison of our LAMNet and LAMNet-large with recent advanced lightweight image SR methods on five benchmark datasets. All the efficiency proxies (Parameter, Flops) are measured for the case of upsampling the image resolution to \(1280\times 720\). The best and second-best results are marked in red and blue colors. '-' means the result is unavailable.}
\label{tbl:quan_comp}
\centering
\footnotesize
\renewcommand{\arraystretch}{1.3}
\begin{tabular}{c|c|c|c|c|c|c|c|c} 
\toprule
\multirow{2}{*}{Scale Factor} & \multirow{2}{*}{Model} & \multirow{2}{*}{Parameters} & \multirow{2}{*}{Flops} & Set5                                             & Set14                                           & BSD100                         & Urban100                       & Manga109                        \\ 
\cline{5-9}
                              &                        &                             &                        & PSNR/SSIM                                        & PSNR/SSIM                                       & PSNR/SSIM                      & PSNR/SSIM                      & PSNR/SSIM                       \\ 
\hline
\multirow{13}{*}{x2}          & EDSR-baseline          & 1370K                       & 316G                   & 37.99/0.9604                                     & 33.57/0.9175                                    & 32.16/0.8994                   & 31.98/0.9272                   & 38.54/0.9769                    \\
                              & IMDN                   & 694K                        & 158.8G                 & 38.00/0.9605                                     & 33.63/0.9177                                    & 32.19/0.8996                   & 32.17/0.9283                   & 38.88/0.9774                    \\
                              & RFDN                   & 534K                        & 123.0G                 & 38.05/0.9606                                     & 33.68/0.9184                                    & 32.16/0.8994                   & 32.12/0.9278                   & 38.88/0.9773                    \\
                              & LatticeNet             & 765K                        & 169.5G                 & 38.15/0.9610                                     & 33.78/0.9193                                    & 32.25/0.9005                   & 32.43/0.9302                   & -/-                             \\
                              & SMSR                   & 985K                        & 351.5G                 & 38.00/0.9601                                     & 33.64/0.9179                                    & 32.17/0.8990                   & 32.19/0.9284                   & 38.76/0.9771                    \\ 
\cdashline{2-9}
                              & ESRT                   & 751K                        &                        & 38.03/0.9600                                     & 33.75/0.9184                                    & 32.25/0.9001                   & 32.58/0.9318                   & 39.12/0.9774                    \\
                              & Omni-SR                & 772K                        & 194.5G                 & 38.22/\textcolor{blue}{0.9613}                   & 33.98/0.9210                                    & 32.36/\textcolor{blue}{0.9020} & \textcolor{blue}{33.05/0.9363} & 39.28/\textcolor{blue}{0.9784}  \\
                              & DLGSANet-light         & 745K                        & 175.4G                 & 38.20/0.9612                                     & 33.89/0.9203                                    & 32.30/0.9012                   & 32.94/0.9355                   & 39.29/0.9780                    \\
                              & \textbf{LAMNet(ours)}                 & 828K                        & 185G                   & 38.22/\textcolor{blue}{0.9613}                   & \textcolor{blue}{34.00}/0.9208                  & 32.35/0.9019                   & 33.03/0.9359                   & \textcolor{blue}{39.33}/0.9782  \\ 
\cdashline{2-9}
                              & SwinIR-light           & 910K                        & 244G                   & 38.14/0.9611                                     & 33.86/0.9206                                    & 32.31/0.9012                   & 32.76/0.9340                   & 39.12/0.9783                    \\
                              & SRFormer-light         & 853K                        & 236G                   & \textcolor{blue}{38.23}/\textcolor{blue}{0.9613} & 33.94/0.9209                                    & \textcolor{blue}{32.36}/0.9019 & 32.91/0.9353                   & 39.28/\textcolor{red}{0.9785}   \\
                              & HiT-SNG                & 1013K                       & 213.9G                 & 38.21/0.9612                                     & \textcolor{blue}{34.00}/\textcolor{red}{0.9217} & 32.35/\textcolor{blue}{0.9020} & 33.01/0.9360                   & 39.32/0.9782                    \\
                              & \textbf{LAMNet-large(ours)}           & 1024K                       & 229G                   & \textcolor{red}{38.27/0.9615}                    & \textcolor{red}{34.07}/\textcolor{blue}{0.9214} & \textcolor{red}{32.38/0.9023}  & \textcolor{red}{33.16/0.9373}  & \textcolor{red}{39.38/0.9785}   \\ 
\hline
\multirow{13}{*}{x3}          & EDSR-baseline          & 1555K                       & 160G                   & 34.37/0.9270                                     & 30.28/0.8417                                    & 29.09/0.8052                   & 28.15/0.8527                   & 33.45/0.9439                    \\
                              & IMDN                   & 703K                        & 71.5G                  & 34.36/0.9270                                     & 30.32/0.8417                                    & 29.09/0.8046                   & 28.17/0.8519                   & 33.61/0.9445                    \\
                              & RFDN                   & 541K                        & 55.4G                  & 34.41/0.9273                                     & 30.34/0.8420                                    & 29.09/0.8050                   & 28.21/0.8525                   & 33.67/0.9449                    \\
                              & LatticeNet             & 765K                        & 76.3G                  & 34.40/0.9272                                     & 30.32/0.8416                                    & 29.10/0.8049                   & 28.19/0.8513                   & -/-                             \\
                              & SMSR                   & 993K                        & 156.8G                 & 34.40/0.9270                                     & 30.33/0.8412                                    & 29.10/0.8050                   & 28.25/0.8536                   & 33.68/0.9445                    \\ 
\cdashline{2-9}
                              & ESRT                   & 751K                        & -                      & 34.42/0.9268                                     & 30.43/0.8433                                    & 29.15/0.8063                   & 28.46/0.8574                   & 33.95/0.9455                    \\
                              & Omni-SR                & 780K                        & 88.4G                  & 34.70/0.9294                                     & 30.57/0.8469                                    & \textcolor{blue}{29.28}/0.8094 & 28.84/0.8656                   & 34.22/0.9487                    \\
                              & DLGSANet-light         & 752K                        & 78.2G                  & 34.70/0.9295                                     & 30.58/0.8465                                    & 29.24/0.8089                   & 28.83/0.8653                   & 34.16/0.9483                    \\
                              & \textbf{LAMNet(ours)}                 & 837K                        & 83.11G                 & 34.71/0.9295                                     & \textcolor{blue}{30.63}/0.8472                  & \textcolor{blue}{29.28}/0.8097 & 28.90/0.8668                   & 34.36/0.9491                    \\ 
\cdashline{2-9}
                              & SwinIR-light           & 918K                        & 111G                   & 34.62/0.9289                                     & 30.54/0.8463                                    & 29.20/0.8082                   & 28.66/0.8624                   & 33.98/0.9478                    \\
                              & SRFormer-light         & 861K                        & 105G                   & 34.67/0.9296                                     & 30.57/0.8469                                    & 29.26/0.8099                   & 28.81/0.8655                   & 34.19/0.9489                    \\
                              & HiT-SNG                & 1021K                       & 99.5G                  & \textcolor{blue}{34.74/0.9297}                   & 30.62/\textcolor{blue}{0.8474}                  & 29.26/\textcolor{blue}{0.8100} & \textcolor{blue}{28.91/0.8671} & \textcolor{blue}{34.38/0.9495}  \\
                              & \textbf{LAMNet-large(ours)}           & 1033K                       & 103G                   & \textcolor{red}{34.75/0.9299}                    & \textcolor{red}{30.65/0.8478}                   & \textcolor{red}{29.31/0.8107}  & \textcolor{red}{29.03/0.8692}  & \textcolor{red}{34.44/0.9497}   \\ 
\hline
\multirow{13}{*}{x4}          & EDSR-baseline          & 1518K                       & 114G                   & 32.09/0.8938                                     & 28.58/0.7813                                    & 27.57/0.7357                   & 26.04/0.7849                   & 30.35/0.9067                    \\
                              & IMDN                   & 715K                        & 40.9G                  & 32.21/0.8948                                     & 28.58/0.7811                                    & 27.56/0.7353                   & 26.04/0.7838                   & 30.45/0.9075                    \\
                              & RFDN                   & 550K                        & 31.6G                  & 32.24/0.8952                                     & 28.61/0.7819                                    & 27.57/0.7360                   & 26.11/0.7858                   & 30.58/0.9089                    \\
                              & LatticeNet             & 777K                        & 43.6G                  & 32.18/0.8943                                     & 28.61/0.7812                                    & 27.57/0.7355                   & 26.14/0.7844                   & -/-                             \\
                              & SMSR                   & 1006K                       & 89.1G                  & 32.12/0.8932                                     & 28.55/0.7808                                    & 27.55/0.7351                   & 26.11/0.7868                   & 30.54/0.9085                    \\ 
\cdashline{2-9}
                              & ESRT                   & 751K                        & -                      & 32.19/0.8947                                     & 28.69/0.7833                                    & 27.69/0.7379                   & 26.39/0.7962                   & 30.75/0.9100                    \\
                              & Omni-SR                & 792K                        & 50.9G                  & 32.49/0.8988                                     & 28.78/0.7859                                    & 27.71/0.7415                   & 26.64/0.8018                   & 31.02/0.9151                    \\
                              & DLGSANet-light         & 761K                        & 44.8G                  & 32.54/\textcolor{blue}{0.8993}                   & 28.84/0.7871                                    & 27.73/0.7415                   & 26.66/0.8033                   & 31.13/0.9161                    \\
                              & \textbf{LAMNet(ours)}                 & 849K                        & 47.5G                  & 32.51/0.8983                                     & \textcolor{blue}{28.87/0.7880}                  & \textcolor{blue}{27.75/0.7427} & 26.72/0.8048                   & \textcolor{blue}{31.26}/0.9169  \\ 
\cdashline{2-9}
                              & SwinIR-light           & 897K                        & 65.2G                  & 32.44/0.8976                                     & 28.77/0.7858                                    & 27.69/0.7406                   & 26.47/0.7980                   & 30.92/0.9151                    \\
                              & SRFormer-light         & 873K                        & 62.8G                  & 32.51/0.8988                                     & 28.82/0.7872                                    & 27.73/0.7422                   & 26.67/0.8032                   & 31.17/0.9165                    \\
                              & HiT-SNG                & 1032K                       & 57.7G                  & \textcolor{blue}{32.55}/0.8991                   & 28.83/0.7873                                    & 27.74/0.7426                   & \textcolor{blue}{26.75/0.8053} & 31.24/\textcolor{blue}{0.9176}  \\
                              & \textbf{LAMNet-large(ours)}           & 1045K                       & 58.4G                  & \textcolor{red}{32.60/0.9000}                    & \textcolor{red}{28.87/0.7886}                   & \textcolor{red}{27.77/0.7434}  & \textcolor{red}{26.82/0.8078}  & \textcolor{red}{31.33/0.9183}   \\
\bottomrule
\end{tabular}
\end{table*}

\subsection{Benchmarking}
\begin{figure*}[ht]
    \centering
    \includegraphics[width=\textwidth]{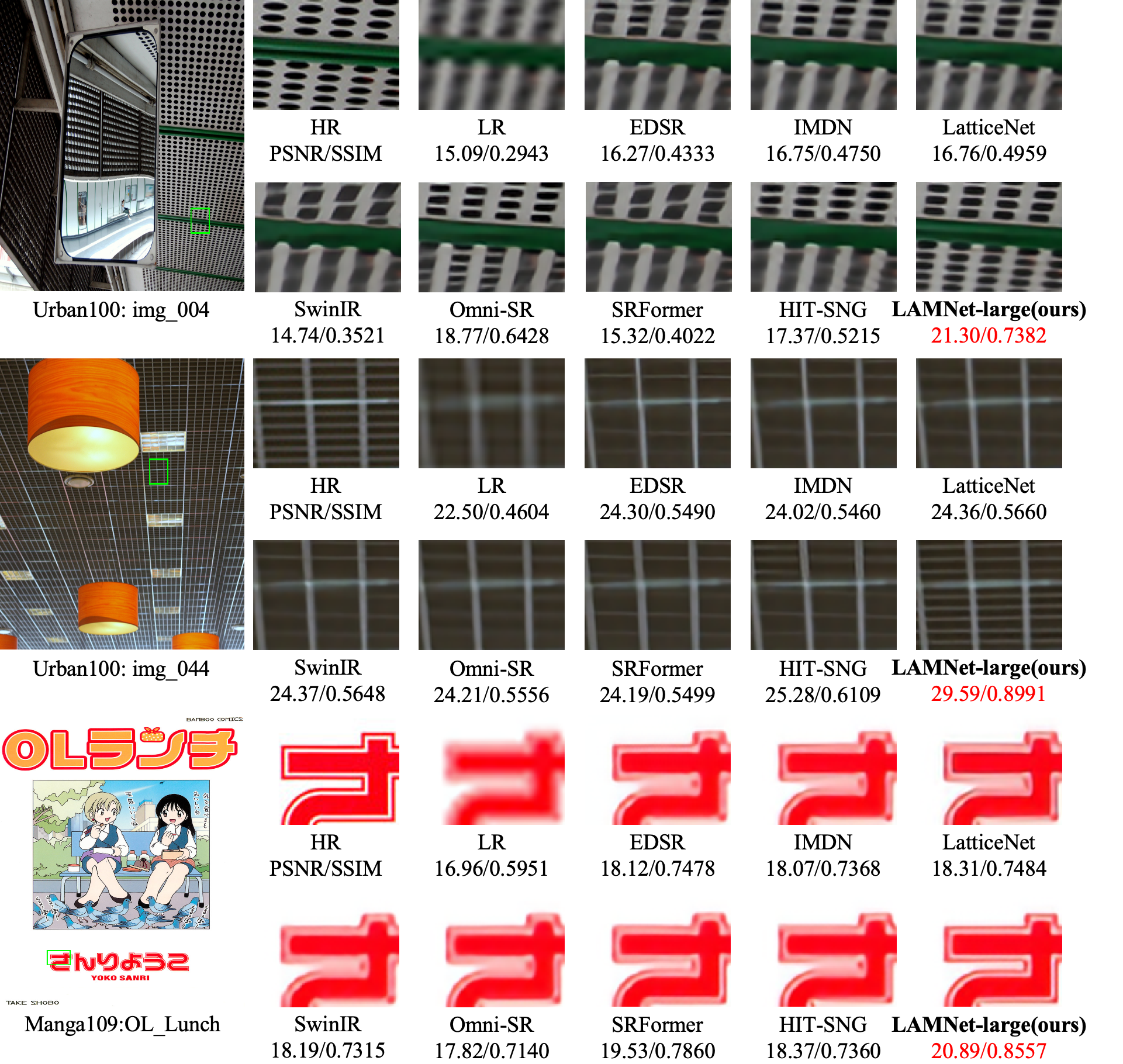}
    \caption{Visual comparisons on Urban100 and Manga109 with scale factor 4.}
    \label{fig:qualitive}
\end{figure*}

\textbf{Datasets and Metrics.} Following previous works, we utilize the DIV2K dataset from the NTIRE 2017 SISR track \cite{conf/cvpr/TimofteAG0ZLSKN17} for our training data. This dataset comprises 1000 HR images, with 800 allocated for model training. To assess our model's effectiveness, we employed five widely used benchmark datasets: Set5 \cite{conf/bmvc/BevilacquaRGA12}, Set14 \cite{conf/cas/ZeydeEP10}, BSDS100 \cite{conf/iccv/MartinFTM01}, Urban100 \cite{conf/cvpr/HuangSA15}, and Manga109 \cite{journals/mta/MatsuiIAFOYA17}. The LR images in these datasets were generated through bicubic downsampling of the original high-resolution images. We evaluated the quality of the reconstructed images using the Peak Signal-to-Noise Ratio (PSNR) and Structural Similarity Index (SSIM) metrics, which were calculated on the Y channel in the YCbCr color space.

\textbf{Implementation Details.} Our proposed LAMNet architecture consists of 4 stacked LAMs. We configured the intermediate feature channels to enhance the model's expressive power to 64. Each LAM module comprises 4 ULMs. For the FSA operation within ULM, we divide the channels into four groups to boost the model's representation capabilities and configure the stride as \([1, 2, 4]\) and the corresponding steps as \([3, 2, 1]\). With this setup, the combined horizontal and vertical FSA achieves an extensive receptive field of \(23\times 23\). When the numbers of the LAM, the ULM, and the feature channel are set to 5, 6, and 64, respectively, we refer to the DLGSANet as DLGSANet-large. During training, \(64\times 64\) patches are cropped from LR images and corresponding patches from HR images. We train the model using the \(L_1\) loss and Adam optimizer for 500k iterations, starting with an initial learning rate of \(1 \times 10^{-3}\) and multiplying with 0.5 after \(\{250, 400, 450, 475\}\)-th epoch for \(2\times\) task. For \(3\times\) and \(4\times\) super-resolution tasks, we initialize the parameters using those from the \(2\times\) task and reduce the total number of training iterations by half. We also randomly utilize 90°, 180°, and 270° rotations and horizontal flips for data augmentation during model training. Additionally, the computational complexity (FLOPs) and runtime of each method are measured based on SR images with a spatial resolution of \(1280 \times 720\).

In this section, we evaluate our newly developed model, LAMNet and LAMNet-large, against leading lightweight models at various SR scale factors, specifically \(2\times\), \(3\times\), and \(4\times\), to evaluate the model's efficacy. The comparison encompasses state-of-the-art efficient SR methods, including CNN-based algorithms like EDSR-baseline\cite{conf/cvpr/LimSKNL17}, IMDN\cite{conf/mm/HuiGYW19}, RFDN\cite{conf/eccv/LiuTW20}, LatticeNet\cite{conf/eccv/LuoXZQLF20}, and SMSR\cite{conf/cvpr/WangDWYLAG21}, as well as Transformer-based methods such as ESRT\cite{conf/cvpr/Lu0LHZZ22}, SwinIR-Light\cite{conf/iccvw/LiangCSZGT21}, OMIN-SR\cite{conf/cvpr/WangCNLL23}, DLGSANet-light\cite{conf/iccv/li2023},SRFormer-Light\cite{conf/iccv/ZhouLGBCH23}, and HIT-SNG\cite{journals/corr/abs-2407-05878}. Our assessment employs quantitative, qualitative, and computational cost analysis.

\textbf{Quantitative Comparisons.} Table \ref{tbl:quan_comp} highlights the strong performance of our proposed LAMNet and LAMNet-Large across all datasets. We group the methods into three categories using dashed lines: the first category represents traditional CNNs, while the second and third are Transformer-based networks, distinguished by their computational complexity. From the table, we observe that Transformer-based methods generally achieve better results with similar model sizes. Compared to other Transformer models, our small-scale LAMNet delivers results on par with or even better than the lightweight OMNI-SR across most datasets, while operating with fewer FLOPs. On the Manga100 dataset, for 3x and 4x tasks, our method improves PSNR by 0.12 dB and 0.24 dB, respectively, representing a notable gain for lightweight tasks. Although LAMNet has more parameters, as discussed in Section \ref{sec:cost_anaysis}, this only slightly increases memory usage. Additionally, the larger version, LAMNet-Large, significantly outperforms the previous lightweight model HIT-SNG, with comparable parameters and FLOPs, achieving over a 0.1dB gain on all scales of the Urban100 dataset.

\begin{figure}[!t]
    \centering
    \includegraphics[width=\columnwidth]{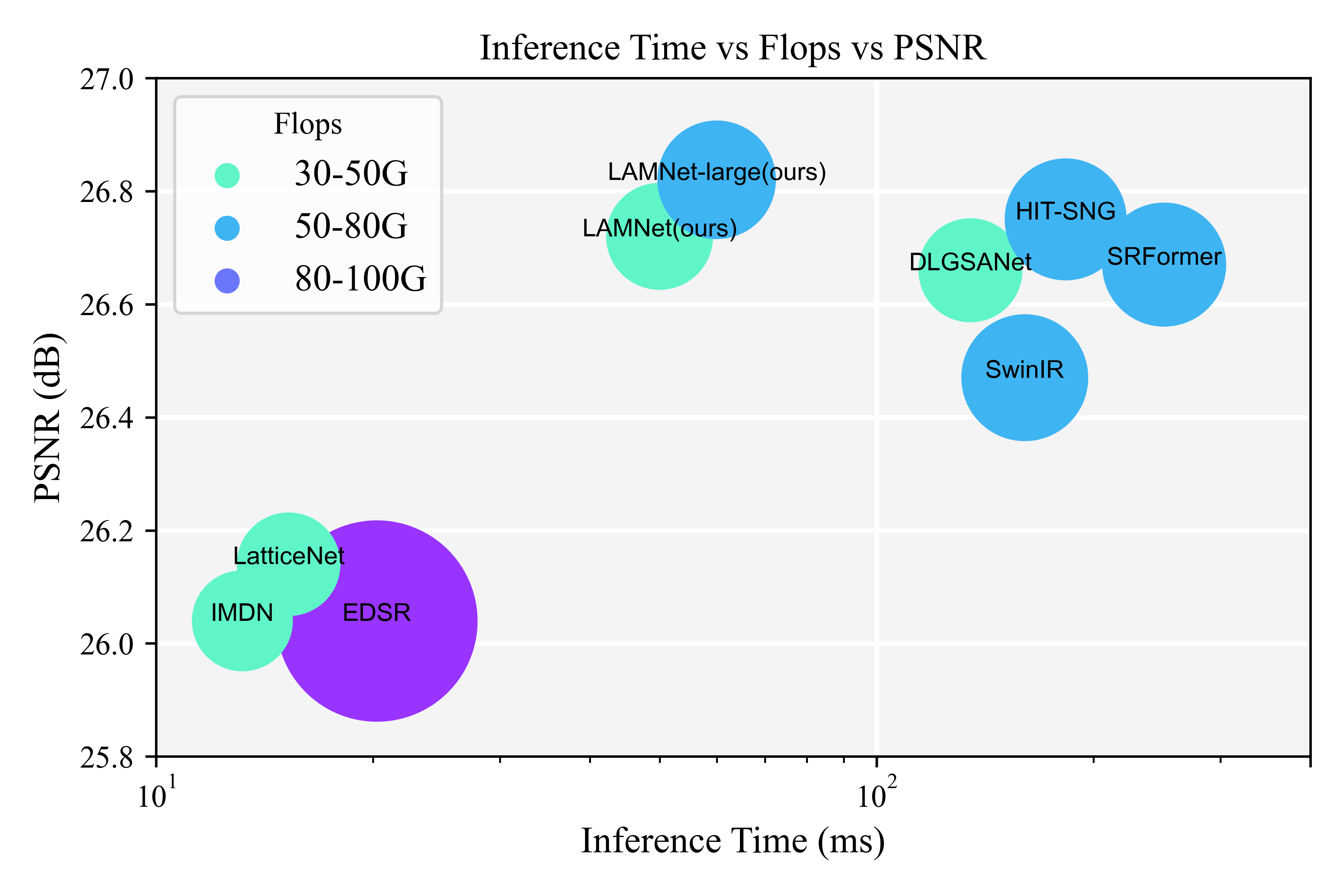}
    \caption{Results are achieved on Urban100 for ×4 SR. LAMNet
attains superior performance while requiring lower computational costs
and incurring lower inference latency.}
    \label{fig:speed}
\end{figure}

\begin{figure}[!t]
    \centering
    \includegraphics[width=\columnwidth]{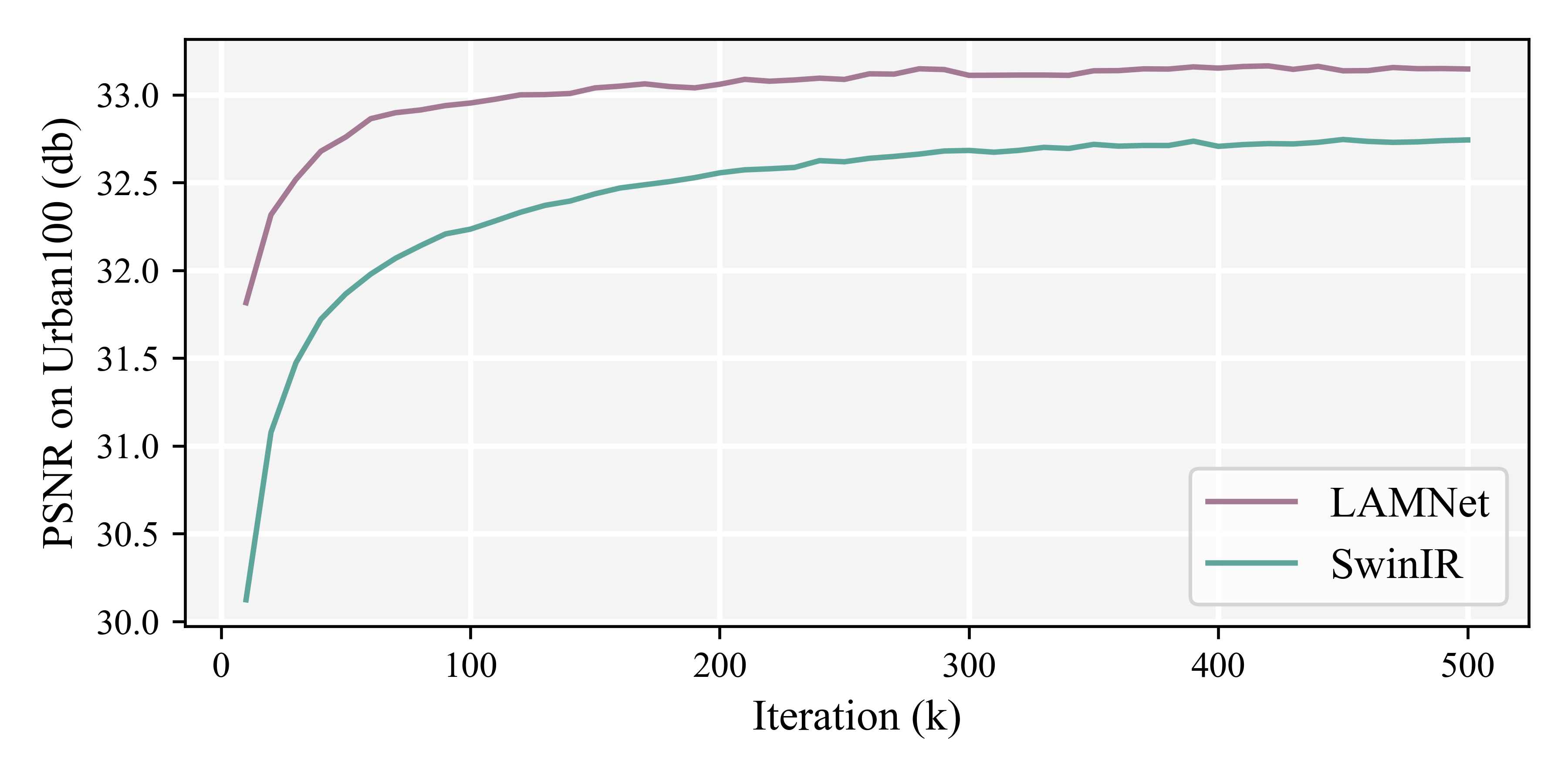}
    \caption{Comparison of the convergence rates between SwinIR-light and our proposed LAMNet-large for the \(4\times\) super-resolution task on the Urban100 dataset.}
    \label{fig:training_curve}
\end{figure}

\textbf{Qualitative Comparisons.} Figure \ref{fig:qualitive} presents qualitative comparisons of various models in challenging super-resolution scenarios. Current SR methods often focus on reconstructing fine, repetitive textures, as seen in scenes like img\_004 and img\_044 from the Urban100 dataset, which depend on surrounding pattern features. Zooming in reveals that previous SR methods produce blurry textures and artifacts, largely due to their limited receptive field and disregard for local texture details. In contrast, our method effectively reconstructs textures by leveraging FSA's large-kernel receptive field and focal mechanism. Beyond texture reconstruction, addressing text adhesion is another crucial challenge in SR tasks, as illustrated in the OL\_Lunch scene from the Manga109 dataset. LAMNet effectively removes the adhesion between strokes, restoring the text’s structure, while other SR methods often struggle with this, resulting in illegible text.

\textbf{Model complexity and Latency.} To clearly illustrate how LAMNet integrates the efficiency of convolution with the performance advantages of Transformer architectures, Figure \ref{fig:speed} compares the inference times of different models on \(4\times\) super-resolution tasks. The x-axis represents inference time, while the y-axis reflects PSNR on the Urban100 dataset. The circle radii represent the relative computational complexity of each model. We observe that convolution-based methods occupy the lower-left region, indicating faster inference but weaker performance, while Transformer-based methods with SA are in the upper-right, showing better performance but longer inference times. Our LAMNet is positioned in the upper-middle, outperforming existing Transformer-based methods in performance while nearing the inference efficiency of convolutional models, achieving a 2 to 3 times speed-up. This is because LAMNet replaces the costly local SA in the Transformer architecture with a more efficient linear dynamic convolution, utilizing the fast inference properties of convolution for acceleration.

Additionally, Figure \ref{fig:training_curve} presents the training curves of our proposed LAMNet-large and SwinIR-light, both with similar computational complexity. We observed that LAMNet achieved better convergence early in the training process, without requiring the extensive iterative training typically needed for SA-based Transformers.

\subsection{Ablation Study}
\begin{table*}
\centering
\small
\caption{We performed ablation experiments under LAMNet’s framework and training settings, testing on the Urban100 and Manga109 datasets. The ablation study includes four groups of experiments: FSA window settings, CSM, IEM, and DGFN. LAMNet's settings are highlighted in bold, and the best results in each group are marked in \textcolor{red}{red}.}
\label{tbl:ablation}
\renewcommand{\arraystretch}{1.3}
\begin{tabular}{c|c|c|c|c|c} 
\toprule
\multirow{2}{*}{Group} & \multirow{2}{*}{Strategies}                   & \multirow{2}{*}{Parameters} & \multirow{2}{*}{Flops} & Urban100                      & Manga109                       \\ 
\cline{5-6}
                       &                                               &                             &                        & PSNR/SSIM                     & PSNR/SSIM                      \\ 
\hline
\multirow{4}{*}{1}     & \textbf{Stride=[1,2,4], Step=[3,2,1], Win=23} & 828K                        & 185G                   & \textcolor{red}{33.03/0.9359} & \textcolor{red}{39.33/0.9782}  \\
                       & Stride=[1,2], Step=[3,3], Win=19              & 822K                        & 184G                   & 32.92/0.9352                  & 39.28/0.9778                   \\
                       & Stride=[1], Step=[6], Win=13                  & 813K                        & 182G                   & 32.88/0.9350                  & 39.24/0.9775                   \\
                       & FSA\(\rightarrow\)MHDLSA,Win=7                              & 839K                        & 186G                   & 32.01/0.9270                  & 38.57/0.9769                   \\ 
\hdashline
\multirow{3}{*}{2}     & \textbf{CSM}                                  & 828K                        & 185G                   & \textcolor{red}{33.03/0.9359} & \textcolor{red}{39.33/0.9782}  \\
                       & CSM\(\rightarrow\)Linear                                    & 828K                        & 185G                   & 32.95/0.9356                  & 39.26/0.9780                   \\
                       & CSM\(\rightarrow\)None                                      & 803K                        & 180G                   & 32.98/0.9358                  & 39.27/0.9780                   \\ 
\hdashline
\multirow{2}{*}{3}     & \textbf{IEM}                                  & 828K                        & 185G                   & \textcolor{red}{33.03/0.9359} & \textcolor{red}{39.33/0.9782}  \\
                       & IEM\(\rightarrow\)None                                      & 828K                        & 185G                   & 32.99/0.9358                  & 39.29/\textcolor{red}{0.9782}  \\ 
\hdashline
\multirow{3}{*}{4}     & \textbf{DGFN}                                 & 828K                        & 185G                   & \textcolor{red}{33.03/0.9359} & \textcolor{red}{39.33/0.9782}  \\
                       & DGFN\(\rightarrow\)GDFN                                     & 760K                        & 170G                   & 32.90/0.9351                  & 39.22/0.9778                   \\
                       & DGFN\(\rightarrow\)FFN                                      & 798K                        & 179G                   & 32.82/0.9344                  & 39.18/0.9778                   \\
\bottomrule
\end{tabular}
\end{table*}

In Table \ref{tbl:ablation}, we examine the effectiveness of each proposed module under LAMNet’s framework and training settings. Note that all models are trained by replacing only the corresponding modules without modifying other settings to ensure a fair comparison.

\begin{figure}[!t]
    \centering
    \includegraphics[width=\columnwidth]{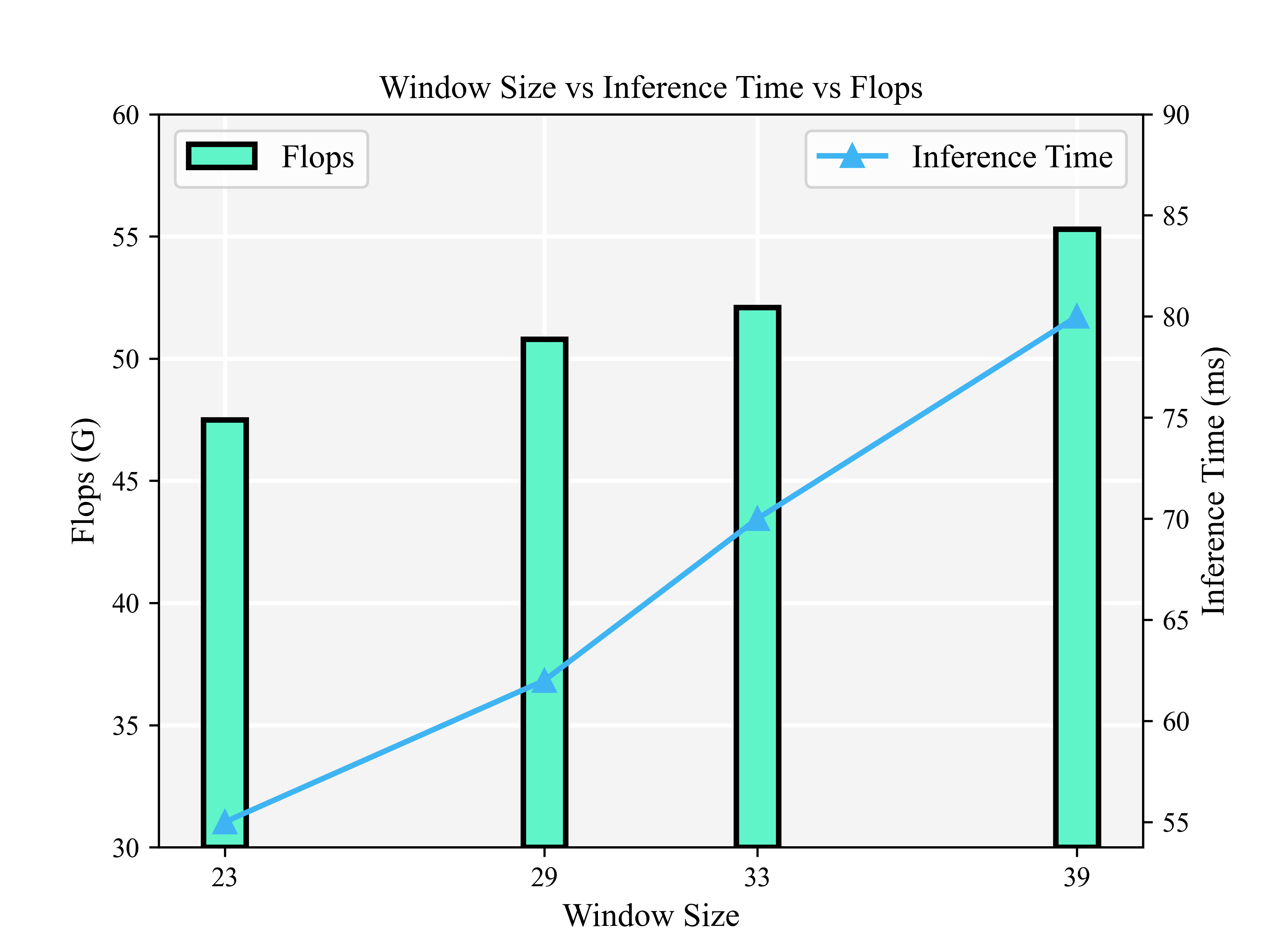}
    \caption{Model's FLOPs and inference time with respect to the window size.}
    \label{fig:win_size_speed}
\end{figure}

\textbf{The effectiveness of FSA.} To evaluate the impact of FSA, we conduct experiments with different window settings to assess the effectiveness of the Focal strategy by gradually reducing the window stride until the Focal operation is eliminated in Group 1. We perform two ablation experiments with stride settings of \([1,2]\) and \([1]\), where the window size decreased from 23 to 19 and finally to 13. As the window size decreased, we observe that the parameters and computational complexity remained almost unchanged, but the overall performance drops significantly. This is because the larger window design allows the model to capture distant similar textures that guide the reconstruction of the current region. Figure \ref{fig:win_size_speed} illustrates the impact of increasing window sizes on both model computational complexity and inference time. While the linearization approach mitigates the growth in computational cost as the window expands, inference time still rises sharply, primarily due to memory access constraints inherent in the linear window design. Additionally, the small patch size used during training further restricts the potential for unlimited window expansion. Furthermore, we replace the linear FSA with the MHDLSA module from DLGSANet. The experimental results reveal a substantial performance degradation following the replacement, indicating that the FSA is better aligned with the LAMNet framework.

\begin{figure}[!t]
    \centering
    \includegraphics[width=0.4\textwidth]{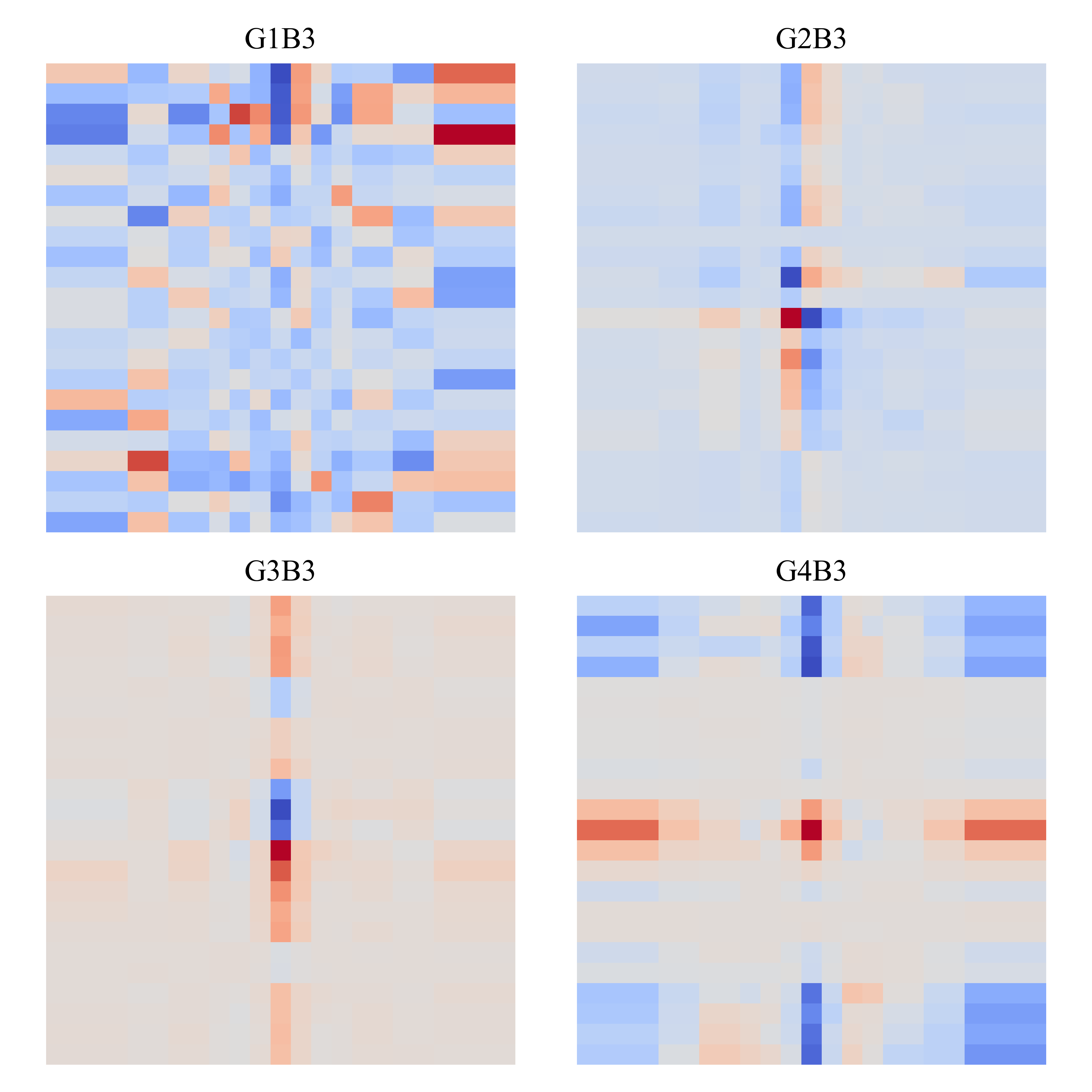}
    \caption{Dynamic weight matrices at the same relative depth across different groups of the model.}
    \label{fig:visual_kernel}
\end{figure}

Figure \ref{fig:visual_kernel} presents a visualization of ULM's dynamic convolution kernel weights for the FSA operation across different layers. These weights are extracted from the same depth within various groups. Firstly, the weight matrices of the dynamic convolution kernels exhibit varying structures, as their rank is greater than 1, which enhances the model’s adaptability and capacity for representation compared to traditional separable convolution. Secondly, the differences in weights across groups enable the model to capture a diverse range of features. In shallower layers, dense matrices aggregate sufficient information, whereas in deeper layers, sparse matrices focus on aggregating finer, more detailed information.

\textbf{The effectiveness of CSM.} In the channel branch, we replace the CSM with a simple linear layer or remove it entirely to validate its role in enhancing channel information. As shown in the second group of experiments in Table \ref{tbl:ablation}, using the linear layer results in a significant performance drop despite a similar computational cost. This can be attributed to the linear layer’s limitation of only combining channel features linearly without introducing nonlinearity or feature selection. Similarly, removing the module also degrades model performance, confirming that the additional channel mechanism in the channel branch benefits the model.

\textbf{The effectiveness of IEM.} We employ the IEM to facilitate feature exchange between the channel and spatial branches, thereby mitigating the limitations of each branch in extracting features. Convolution-based \cite{conf/iccv/0014ZGKY023} and SA-based \cite{journals/corr/abs-2407-05878} feature exchange methods typically introduce high computational complexity, while IEM aims to achieve efficient exchange with minimal computational overhead. Since IEM only involves a few dot-product operations, the computational cost remains minimal. The third group of experiments in Table \ref{tbl:ablation} demonstrates that while removing this module does not significantly alter the model’s size, it degrades the model’s performance by hindering information exchange between the branches.

\begin{figure}[!t]
    \centering
    \includegraphics[width=\columnwidth]{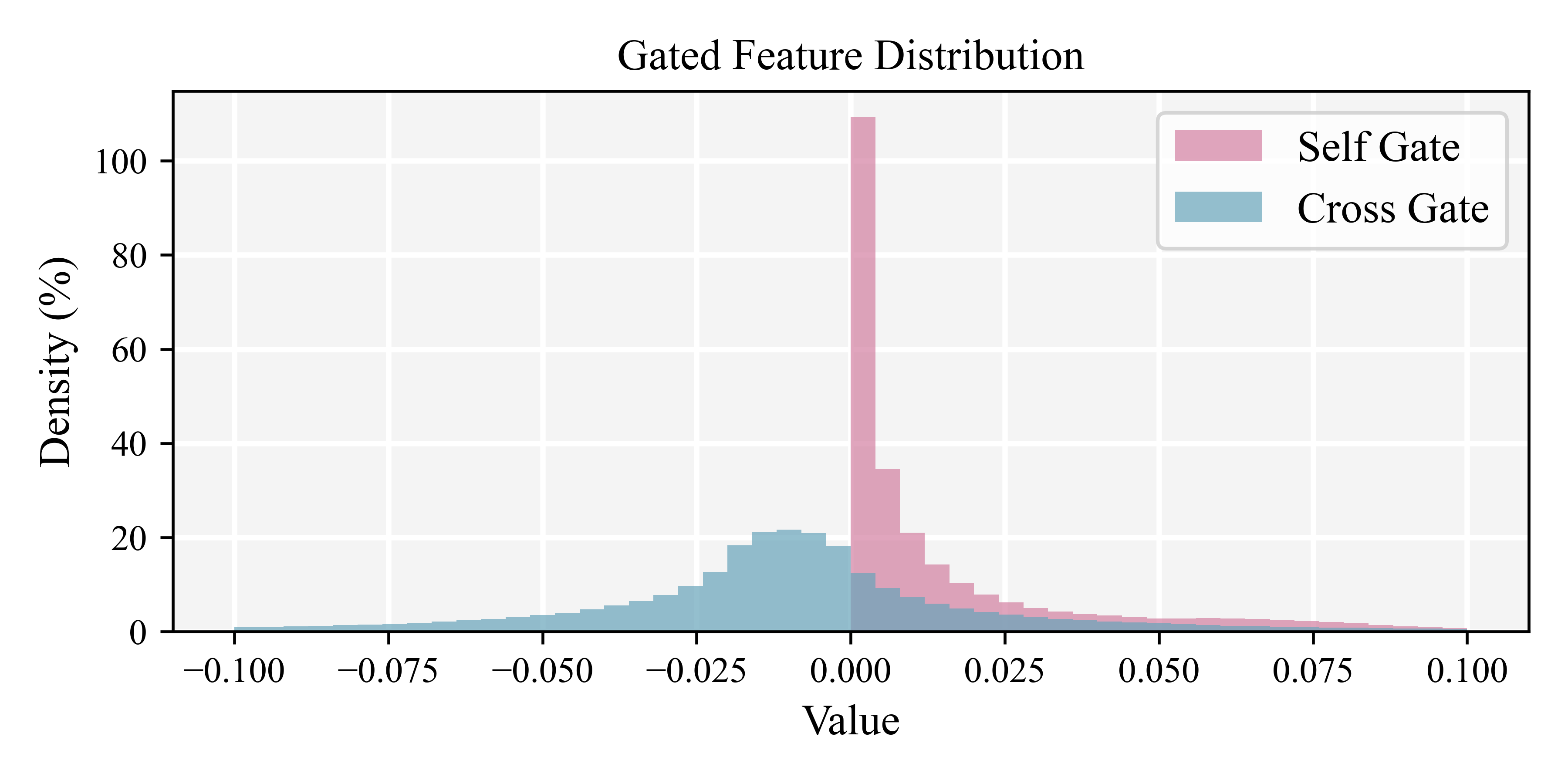}
    \caption{Feature distributions following the Self-Gate and Cross-Gate in the DGFN.}
    \label{fig:gated_feat}
\end{figure}

\textbf{The effectiveness of DGFN.} The Feed-Forward Network (FFN) is a critical component of Transformer models and is typically applied after token information exchange in NLP tasks, where it introduces non-linearity to high-dimensional token channels, allowing for the exploration of more complex relationships. As shown in the fourth group in Table \ref{tbl:ablation}, our framework experiences a significant performance drop when using a traditional FFN, mainly due to its limited capacity for spatial information exploration. Nonetheless, thanks to the superior design of our overall architecture, our model still outperforms the FFN-based SwinIR-light, even with fewer FLOPs. Similarly, some methods \cite{conf/cvpr/ZamirA0HK022, conf/iccv/ZhouLGBCH23} have recognized the importance of spatial dimensions in FFN and have proposed solutions to address this, such as the Gated-Dconv feed-forward network (GDFN) in Restormer \cite{conf/cvpr/ZamirA0HK022}, which mitigates this shortcoming. However, the spatial-gate approach conflicts with the FFN’s original goal of modeling high-dimensional spaces. According to the results from the Urban100 and Manga109 datasets shown in Table \ref{tbl:ablation}, GDFN introduces a self-gate that preserves dimensional information lost during cross-gate operations, allowing the model to maintain the ability to explore channel dimensions, leading to a performance gain of over 0.1dB. Additionally, Figure \ref{fig:gated_feat} illustrates the features produced by the self-gate and cross-gate mechanisms within a specific layer of the DGFN. The self-gate emphasizes positive-valued features, whereas the cross-gate exhibits a more uniform feature distribution, though with a slight mean shift. Retaining both types of feature distributions, rather than following the GDFN approach that only retains cross-gate features, contributes positively to the model's ability to capture more complex channel relationships.

\subsection{Limitations}
Our proposed LAMNet leverages dynamic convolution to effectively approximate the adaptive aggregation capability of SA while matching the superior performance of Transformers and the inference efficiency of convolutional networks. However, there remains a noticeable gap in inference time compared to convolutional networks of the same scale. This discrepancy is partly due to using LayerNorm and the GELU activation function in the Transformer framework. Additionally, the dynamic convolution in LAMNet has not undergone hardware-specific optimizations like those implemented in DCNv4 \cite{conf/cvpr/XiongYuwen24}, indicating that there is still room for further optimization.
\section{Conclusion}\label{sec:conclusion}
In this paper, we offer new insights into replacing the original local SA mechanism in the Transformer framework with a convolution-based linear adaptive Mixer. By employing a simple dual-branch structure combined with IEM, we enhance the Token Mixer's ability to extract diverse features with minimal overhead. Moreover, we discovered that the gate operations designed to improve the spatial extraction capacity of FFN can impair its channel dimension information. To address this limitation, we utilize a straightforward self-gate mechanism to preserve the dimensional information of gated features. Consequently, we fully replace the computationally expensive SA operation with a convolution-based approach. Furthermore, extensive experiments demonstrate that our LAMNet effectively combines the superior performance of Transformer architectures with the inference efficiency of convolutional neural networks.

{
    \small
    \bibliographystyle{ieeenat_fullname}
    \bibliography{main}
}

\end{document}